# Automating the Surveillance of Mosquito Vectors from Trapped Specimens Using Computer Vision Techniques

Mona Minakshi
University of South Florida
Tampa, Florida, USA.
mona6@mail.usf.edu

Pratool Bharti
Northern Illinois University
Dekalb, Illinois, USA.
pbharti@niu.edu

Willie B. McClinton III
University of South Florida
Tampa, Florida, USA.
wmcclinton@mail.usf.edu

Jamshidbek Mirzakhalov
University of South Florida
Tampa, Florida, USA.
mirzakhalov@mail.usf.edu

Ryan M. Carney
University of South Florida
Tampa, Florida, USA.
ryancarney@usf.edu

Sriram Chellappan
University of South Florida
Tampa, Florida, USA.
sriramc@usf.edu

## ABSTRACT

Among all animals, mosquitoes are responsible for the most deaths worldwide. Interestingly, not all types of mosquitoes spread diseases, but rather, a select few alone are competent enough to do so. In the case of any disease outbreak, an important first step is surveillance of *vectors* (i.e., those mosquitoes capable of spreading diseases). To do this today, public health workers lay several mosquito traps in the area of interest. Hundreds of mosquitoes will get trapped. Naturally, among these hundreds, taxonomists have to identify only the vectors to gauge their density. This process today is manual, requires complex expertise/ training, and is based on visual inspection of each trapped specimen under a microscope. It is long, stressful and self-limiting. This paper presents an innovative solution to this problem. Our technique assumes the presence of an embedded camera (similar to those in smart-phones) that can take pictures of trapped mosquitoes. Our techniques proposed here will then process these images to automatically classify the genus and species type. Our CNN model based on Inception-ResNet V2 and Transfer Learning yielded an overall accuracy of 80% in classifying mosquitoes when trained on 25, 867 images of 250 trapped mosquito vector specimens captured via many smart-phone cameras. In particular, the accuracy of our model in classifying *Aedes aegypti* and *Anopheles stephensi* mosquitoes (both of which are especially deadly vectors) is amongst the highest. We also present important lessons learned and practical impact of our techniques in this paper.

## 1 INTRODUCTION

Mosquito-borne diseases like malaria, dengue, West Nile virus and Zika fever are significant public health concerns, with major human and economic cost. For instance, malaria alone is responsible for more than 1 million deaths per year worldwide, with most of them being children. The year 2019 was recorded as the worst for dengue in South East Asia. There are generally no vaccines or cures available for these diseases, and thus prevention relies upon mosquito surveillance and control. This in-turn requires accurate and real-time knowledge on the geographic presence and abundance of mosquito *vectors* (i.e., those that are capable of spreading diseases). This is significant because, while there are close to 4, 500 species of mosquitoes (spread across 34 or so genera) [16], only a select few are competent vectors. Mosquito vectors primarily belong to three genera - *Aedes* (*Ae.*), *Anopheles* (*An.*) and *Culex* (*Cu.*). Within these genera, there are multiple species responsible for transmitting particular diseases. Malaria is spread primarily by *An. gambiae* in Africa and by *An. stephensi* in India. Dengue, yellow fever, chikungunya, and Zika fever are spread primarily by the species *Ae. aegypti. Cu. nigripalpus* is a vector for West Nile and other encephalitis viruses.

Importantly, in areas where mosquito-borne diseases are prevalent, the vectors co-exist with other mosquito types that are not competent enough to spread diseases. For instance in places like Florida, South America, Uganda and Southern India, where mosquito-borne diseases are problematic, more than 50 species of mosquitoes co-exist with the vectors there. As such, it is always paramount to constantly know which (i.e., genus and species) types of mosquitoes are present, and in what densities to learn about vector populations. Across the world, this is done today by public health workers laying many mosquito traps. Hundreds get trapped each day. Subsequently, these mosquitoes are all bought to a lab like facility where taxonomists (with years of experience) visually inspect each and every trapped specimen under a microscope for identifying its genus and species type. The process is the same in Brazil, India, Rwanda, Uganda and the USA, where we have partners. The process takes hours each day, is cognitively very demanding, and stressful to the taxonomists.

### 1.1 Our Overarching Goal

Our overarching goal in this paper is to automate the process of identifying mosquito vectors with minimal manual/ expert intervention. In the simplest case, we envisage a flat surface over which there is a movable camera. As mosquitoes are spread on the flat surface, the camera moves/ tilts and takes multiple pictures of each mosquito specimen (under normal light conditions). Subsequently, the images will be exported to the cloud where they will be processed to identify the genus and species type of each specimen. This information is fed back instantly to public health facilities with appropriate alerts and images via a dashboard if the ones identified are vectors of interest. If enabled in practice, a technique like ours can bring in significant benefits in terms of speed, accuracy and cost-savings for vector surveillance across the globe and particularly in low-income countries where taxonomists are increasingly hard to find, but wherein surveillance of vectors is a critical component to combat disease spread.

## 1.2 Our Technical Contributions

To develop an AI model, we participated in mosquito trapping experiments run by the Hillsborough county in Florida. Our dataset generated was 25, 867 images of 250 specimens trapped in commercial mosquito traps, each of which was frozen and then visually examined by expert taxonomists under a microscope to identify genus and species. We then captured multiple images of each mosquito specimen by emplacing the mosquito on a flat surface with a smartphone camera attached to a movable fixture placed a feet above the mosquito. A total of 6, 807 images were captured by us using combinations of multiple (three) backgrounds, multiple (upto ten) phones and multiple (three) camera orientations. These 6, 807 images were augmented using standard techniques to generate a dataset of 25, 867 images to be used for model training. To the best of our knowledge, we believe such a large scale data-set of mosquito vectors trapped in nature and tagged with genus and species identifiers is unique. In this realm, our technical contributions are the following:

- **Designing Neural Network Architectures:** We design multiple neural network architectures employing the principle of Transfer Learning on the Inception-ResNet V2 (IRV2) architecture for multiple classes of problems - genus classification; species classification; and species classification based on known genus. Each problem is important in different contexts and are elaborated later in the paper.
- **Enabling Explainable AI by visualizing the feature maps of last convolution layer:** In order to evaluate the robustness of our technique, and to provide insights to public health experts, we visualised the feature maps of last convolution layer [30] to highlight those pixels within the image that were most significant in genus/ species classification. We find that our technique is robust in that the pixels that were most important for classification in our AI model were actually those that represent the core anatomical components of a mosquito - thorax, wings, abdomen and legs. These findings proved to be important to gain the confidence of our public health partners as they validated our techniques.
- **Classification Results:** We contextualize our results in this paper, and evaluate accuracy of our techniques in classifying all vectors in our dataset. Very interestingly, the highest accuracies were obtained in classifying *Aedes aegypti* and *Anopheles stephensi*, both of which are among the most competent vectors known to mankind - the former for its potency in spreading Zika fever, dengue, yellow fever and chikungunya; and the latter for malaria. Taxonomists in Uganda, India, Brazil and the USA that we spoke to were encouraged with these results.
- **Contextualizing our AI results with existing knowledge about evolution of mosquito phylogeny:** We discovered that our classification results gleaned from AI techniques actually reflect existing knowledge gleaned from decades of research using visual markers and DNA analyses on phylogeny of mosquitoes. These comparisons, also elaborated upon in this paper towards the end further add much more robustness and validity to our AI techniques, and also their broader impact in orthogonal fields.
- **Scoping out the practical impact of our work:** Upon discussion with ten experts in Taxonomy, Entomology, Public Health (in India, Brazil, Sub-Saharan Africa and USA), we have identified the practical impact of our work, and how to scale it up for global health. These discussions are all presented towards the end of the paper.

The rest of the paper is organized as follows. In Section 2, important work done related to our problem are discussed. In Section 3, we present the generation of image database and noise removal process. Section 4 contains the details of our neural network architectural design to classify mosquitoes. In Section 5, we discuss about our evaluation of our architectures. Section 6 contains the discussions on the practical impact of our contribution. Finally, we conclude the paper in Section 7.

## 2 RELATED WORK

In the context of combating mosquito-borne diseases, corporations like Microsoft, and agencies like NASA, Global Mosquito Alert Consortium and the U.S. military are investing in drone and satellite based technologies to identify mosquito habitats. Organizations like iNaturalist are also encouraging citizens across the world to generate and upload images of animals and insects such as mosquitoes they encounter in nature for large scale data collection. Nevertheless, of interest to this paper is related work on AI technologies for identification of trapped mosquitoes, which we present below.

In [4], a Support Vector Machine based algorithm was designed and evaluated to detect *Ae. aegypti* mosquitoes using a dataset of 40 images taken via a 500x optical zoom camera. In other related work [6] and [8], learning algorithms were designed to detect mosquitoes from other insects like bees and flies, where the images are captured once again using sophisticated digital cameras. In another 2016 paper [27], the authors imaged wings, and more specifically looked at coordinates at intersections of wing veins as a feature to classify genus of mosquitoes. However, only 12 mosquitoes in total were considered for model development, and also, the requirement of imaging wings alone is cumbersome. In another related work in [18], the authors have designed a Support Vector Machine-based algorithm to identify only the species of a mosquito specimen captured via smartphone images. But this work in [18] has limitations. The dataset in [18] included only 303 images spread across nine mosquito species.

In contrast to the above works, our dataset in this paper contains 25, 867 images of nine mosquito species taken with many more types of smartphones, and in multiple backgrounds and orientations. With this larger scale dataset, we design deep neural net

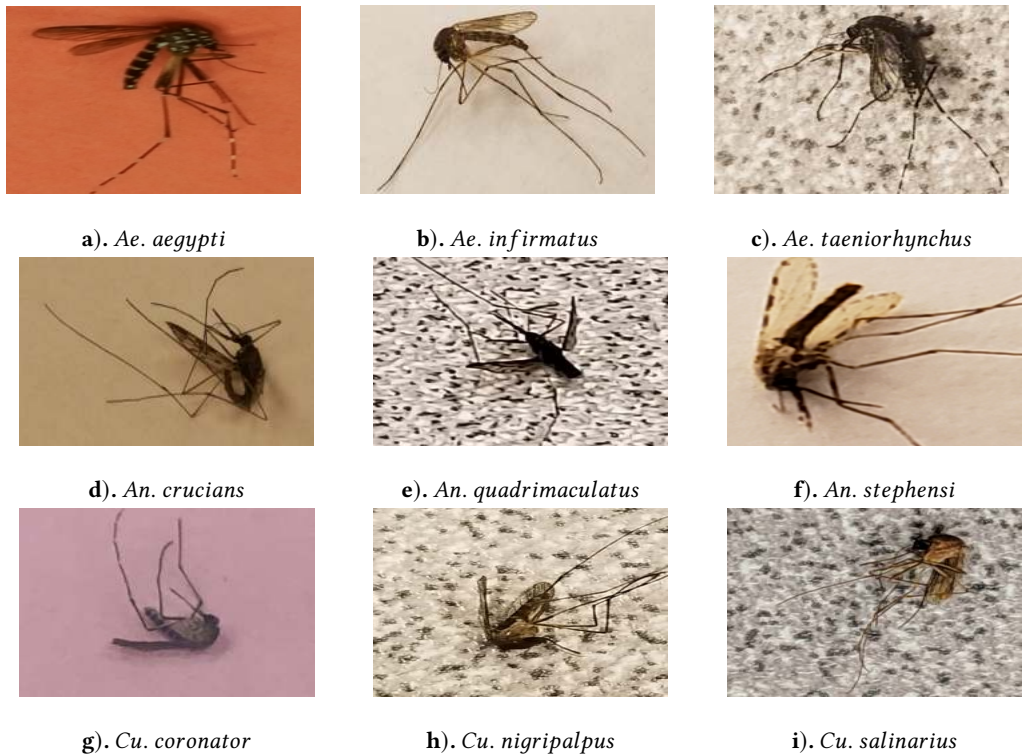

**a)**. *Ae. aegypti*  **b)**. *Ae. infirmatus*  **c)**. *Ae. taeniorhynchus*

**d)**. *An. crucians*  **e)**. *An. quadrimaculatus*  **f)**. *An. stephensi*

**g)**. *Cu. coronator*  **h)**. *Cu. nigripalpus*  **i)**. *Cu. salinarius*

Figure 1: One representative sample in our dataset for each species classified. This figure is best viewed in color.

architectures (which are state-of-the-art) in this paper for three classes of problems a) to identify genus alone; b) to identify species based on knowledge of genus type; and finally c) to directly identify the species type. We also contrast the performance of each architecture and provide contextual relevance to ensuing results. As such, we believe that the work in this paper is significantly advanced, and enables more practical automation for surveillance of vector mosquitoes in nature using image data, compared to existing works in this space.

In parallel, researchers are also looking at acoustic markers to classify mosquitoes. Work here include [3] and [20], wherein the authors use "optical" sensors, and smartphone microphones to record sounds of mosquito flight, and then use features derived from sound for classification of mosquitoes. Ten mosquito species formed the dataset to achieve accuracies in the range of 90% (although, we could not glean information on number of specimens used for model development in [3] and [20]). We do acknowledge that these techniques are certainly innovative and may find acceptance in the future. But public health experts we partnered with were a little skeptical with acoustic techniques, since wing beat frequencies may change with size of the wing as the mosquito ages (but morphology stays very consistent). Furthermore if experts want to manually verify a classification made by an AI algorithm (for example, in the case of a specific vector), they were very unsure about listening to the wing beat frequency manually to confirm assessment. They indicated very clearly that they always prefer one more images for any verification compared to audio files containing wing beats.

As such, these orthogonal techniques may not be practical as of today, unlike image based classification approaches for mosquito surveillance.

Recently, researchers are also exploring ways to identify *larvae* of mosquitoes from citizen generated imagery [21]. Using a dataset of 550 images generated by citizens, an AI framework is designed in [21] to identify only images of mosquito larvae. However, beyond this, there is no method we are aware of that can identify larvae of vectors from larvae of non-vectors, towards smarter surveillance like we attempt in this paper.

## 3 GENERATION OF IMAGE DATABASE AND NOISE REMOVAL

### 3.1 Image Generation

We present a brief overview of our data collection procedure here. Between Fall 2018 and Spring 2019, we visited the Hillsborough county mosquito control board in Florida to lay mosquito traps, collect specimens the next day, and bring them back to the lab for identification. We point out there are close to 50 or different species of mosquitoes in Hillsborough County in Florida. But to keep the problem tractable, without losing the focus on public health, our team decided to only look for vectors of critical interest. From among the hundreds of mosquitoes trapped in our traps, 250 specimens were identified as critically important vectors by our taxonomy partners. These vectors were categorized into three species each belonging the three genera - *Aedes*, *Anopheles* and

Table 1: Relevant Details on our Dataset of Mosquito Species [2]

| Genus | Species | No. of Image Samples | Diseases Spread | Geographical Location |
|---|---|---|---|---|
| Aedes | *aegypti* | 759 | Chikungunya, Dengue, Yellow Fever, Zika | Africa, Asia, North America, South America |
| Aedes | *infirmatus* | 741 | Eastern Equine Encephalitis | North America, South America |
| Aedes | *taeniorhynchus* | 761 | Eastern Equine Encephalitis, West Nile | North America, South America |
| Anopheles | *crucians* | 790 | Malaria | Africa, North America, South America |
| Anopheles | *quadrimaculatus* | 810 | Malaria | North America |
| Anopheles | *stephensi* | 756 | Malaria | Africa, Asia |
| Culex | *coronator* | 712 | St. Louis Encephalitis, West Nile | Africa, North America, South America |
| Culex | *nigripalpus* | 703 | Eastern Equine Encephalitis, St. Louis Encephalitis, West Nile | Africa, North America, South America |
| Culex | *salinarius* | 775 | Eastern Equine Encephalitis, St. Louis Encephalitis, West Nile | Africa, North America, South America |

*Culex*. The dataset is presented in Table 1. Importantly, mosquitoes belonging to the *Aedes*, *Anopheles* and *Culex* genera, and are also the deadliest vectors across the globe.

Subsequently, each specimen was imaged by our team under normal indoor light conditions using ten different smartphones, across three orientations, and on three different backgrounds. To capture each image, the smart-phone was attached to a movable fixture a feet above a flat surface on which the mosquito was placed. The ten phones used for imaging were Samsung Galaxy S8 (3 phones), Samsung Galaxy S9 (2 phones), iPhone 7 (2 phones), iPhone 8 plus (2 phones), and Pixel 3 (1 phone). The three backgrounds were white, pink and a cream colored (flat) tile. As a result of this process, 6807 images in total were captured by our team. Figure 1 illustrates one representative image from our dataset of each of the nine species (spread across three genera) we classify in this paper [1]. Recall that all are vectors.

While our dataset may seem to be relatively large and comprehensive for model development, there is one practical challenge that needs to be overcome. During run-time execution, unseen variances will creep in images taken that will compromise classification. Such variances are numerous and stem from camera hardware, degree of zoom, blurriness, ambient light and more. While it is simply impossible to account for every possible source of variance, we still augment images in our training dataset by zooming in (randomly chosen between 105% and 150%), zooming out (randomly chosen between 75% and 90%), increasing brightness/ contrast (randomly chosen between 1.05 and 1.50) and decreasing brightness/ contrast (randomly chosen between 0.75 and 0.95). These variations are reasonable for our problem/ data collection set-up. As such, the total number of mosquito images generated and augmented for model development and validation was 25, 867.

**A note of gender of mosquitoes trapped:** We point out that all the traps laid to capture mosquitoes were $CO_2$ traps. This meant that the specimens we trapped were all female, since only female mosquitoes seek out a bloodmeal (to provide nutrients for their eggs), and they are attracted to $CO_2$ emanating from the traps, thinking that the $CO_2$ comes from a blood source. Males only feed on plant nectar. As such, our model is trained to only recognize female mosquitoes. But, male and female mosquitoes belonging to the same genus and species are anatomically similar except that the males have a distinct feather like antennae [15]. As such, we are confident that our technique can be easily adapted to identify male mosquitoes as well (not done in this paper though).

---

[1] Among the nine species classified in this paper, eight were trapped in the wild. Only *Anopheles stephensi* mosquitoes were raised in a lab environment in Florida, whose ancestors were originally trapped in India.

## 3.2 Noise Removal

Mosquito images taken from smartphone cameras contain three common types of noise: random noise, fixed pattern noise and banding noise [14]. These types of noise stem from camera hardware, ambient conditions, variations in shutter speed, hand movements etc., and must be removed prior to classification. It is easy to infer that similarities within each anatomical component of a mosquito image should ideally be leveraged for noise removal. But, as we see in Figure 1, the same anatomical component (e.g., legs and wings of a mosquito) appear at multiple locations within a single image. Thus, exploiting purely localized trends in the image for noise removal has limited applicability for our problem. As such, we employ a non-local means denoising technique [1], wherein a mosquito image is smoothed by taking the mean RGB value of all pixels in the image, weighted by how similar these pixels are to any target pixel. This method ensures that similarities in anatomical components within a mosquito are effectively integrated, hence resulting in superior post-filtering clarity, and less loss of detail in the image. We explain below.

Consider a noisy mosquito image $v = \{v(i) \mid i \in I\}$, where $v(i)$ is the RGB value of pixel $i$. Then, the estimated value $NL[v](i)$, for a pixel $i$ in image $v$ after noise removal, is computed as a weighted RGB average of all pixels in the image as,

$$NL[v](i) = \sum_{\forall j} w(i,j)\, v(j),$$

where the weights $w(i,j)$ depend on the similarity between the pixel of interest $i$ and every other pixel $j$ in the image, and satisfy the conditions $0 \leq w(i,j) \leq 1$ and $\sum_{j} w(i,j) = 1$. The similarity between two pixels $i$ and $j$ depends on the similarity of the intensity vectors $\hat{v}(N_i)$ and $\hat{v}(N_j)$, where $\hat{v}(N_i)$ denotes RGB values of entries in a square matrix $N_i$, which in-turn denotes a square neighborhood of fixed size and centered at a pixel $i$. This similarity is measured as as an Euclidean distance, $||\hat{v}(N_i) - \hat{v}(N_j)||^2$. With this method, the goal is to ensure that those pixels within the image that represent the same anatomical component of pixels in the neighborhood $\hat{v}(N_i)$ of any pixel $i$ (chosen for noise removal) will have higher weights on the average, hence smoothing the image more effectively and retaining anatomical consistencies within the image. The weight $w(i,j)$ is computed as

$$w(i,j) = \frac{1}{Z(i)} e^{-\frac{||\hat{v}(N_i) - \hat{v}(N_j)||^2}{h^2}},$$

where $Z(i)$ is normalizing factor given by,

$$Z(i) = \sum_{j} e^{-\frac{||\hat{v}(N_i) - \hat{v}(N_j)||^2}{h^2}},$$

and where $h$ denotes the degree of filtering. For best results, we kept the neighborhood size around a target pixel as [7, 7] and $h$ was set as 10. All images in our dataset were processed in this manner to remove noise before model training.

# 4 OUR DEEP NEURAL NETWORK ARCHITECTURE TO CLASSIFY MOSQUITOES

We now present our neural net architecture to classify mosquitoes from smartphone images. To keep this simple, and easier to explain, we first present our technique to identify only the *genus* of the mosquito from an image. Once we do that, we then present how the technique is adapted for classifying species based on genus, and classifying species directly.

## 4.1 Architecture Rationale and Transfer Learning

The problem of classifying mosquitoes from smartphone images is complex and challenging. Some obvious challenges include need for faster training, reducing computational overhead, and minimizing overfitting. Apart from these, one unique challenge is that the size and position of the actual 'mosquito' within an image is not consistent as we see in Figure 1. Traditionally stacking convolution layers deeper and deeper for learning is computationally expensive and will not give good performance [25]. Furthermore, they cannot effectively handle images, where the object of interest, in this case, a mosquito, can be of any arbitrary size and be at an arbitrary location within the image captured.

To overcome these challenges, we started our model development with the Inception Net architecture [26]. In this architecture, multiple kernel sizes (compared to a single sized kernel) are used at the same layer to better compensate for objects of interest being at arbitrary locations and of arbitrary size in an image. Inception Net architecture enables smart factorization methods to break down larger sized kernels into multiple smaller sized kernels that speed up operation without compromising learning.

However, this itself is not enough for our problem. As the network goes deeper, the gradient of the activation function that propagates through these layers becomes smaller, effectively preventing learning at later layers. To overcome this problem, the Inception Net architecture is combined with another popular neural network architecture called ResNet [12]. Basically, in the ResNet architecture, residual connections are introduced that is a connection from layer $x$ directly to layer $x + n$ (where $n > 1$). In this modification, gradients that are sufficient to improve learning are still maintained at later layers, hence improving learning ability of the architecture, while also speeding up training. To summarize here, the initial architecture we started with for our problem is the Inception-ResNet-V2 (IRV2) [25] architecture ( that combines the Inception and ResNet architectures), which has a total of 782 layers. The IRV2 architecture has been trained on the well known Imagenet dataset [5] that consists of 14 million images categorized across 1000 diverse classes commonly encountered daily, and the architecture achieves very good accuracies in classification for this dataset.

## 4.2 Optimization of Hyperparameters

Now that the rationale of our neural net architecture for mosquito genus identification is clarified, there are several key parameters in the architecture (called as hyperparameters) that we need to optimize. We elucidate these below, following which the final architecture for genus identification will be presented. Also note that since the three problems of interest to this paper (genus, species within genus, and direct species identification) all relate to classifying mosquitoes, the choice of hyperparameters are not going to be too different for each problem. Also, as it standard practice in designing deep neural net architectures, the choice of hyperparameters are finalized through repeated training and validation on the data set. We do not present all details of all hyperparameters that we attempted during training in this paper, but only discuss choices of critical hyperparameters that together gave us the highest accuracies and contextual correctness during training and validation on our mosquito image dataset.

*a). Image Resizing*: To keep the images consistent, we need resizing. Since, for our problem, we collected data from multiple smartphones, we had images of different resolutions ranging from $450 \times 550$ to $2988 \times 5322$ pixels. To bring uniformity, and reduce the image size (for faster training without loss of quality), we resized each input image to $299 \times 299$ pixels irrespective of their original size. Finally, we normalized the RGB value of each pixel of the image by dividing it by 255 before training starts.

*b). Optimizer*: An optimizer is an algorithm which helps converge an architecture during training from an initial value to the optimized one where the loss is minimum. In this study, we have employed Adam (Adaptive Moment Estimation) [17] optimizer algorithm. The idea here is to use adaptive learning rates for weights among layers in the architecture, such that lower rates are assigned to the weights that are getting bigger updates, and higher rates are assigned to weights that are getting smaller updates. Parameters $\beta_1$ and $\beta_2$ here are set as 0.89 and 0.999. Note that $\beta_1$ and $\beta_2$ are the exponential decay rates for the first moment and second-moment estimates respectively.

*c). Loss Function*: We employed the categorical cross entropy loss function in this paper. This function minimizes the divergence of predicted and actual probability function. This is in comparison with other loss functions like focal loss and triplet loss functions that work better when variations in terms of complexities of entities within classes and their inter-variabilities are higher, neither of which is not true for our problem.

*d). Learning Rate*: In this paper, we have utilized Cyclic Learning Rate (CLR) [23] technique. The core idea here is to let the learning rate vary within a range of values, instead of pre-defining a linearly or an exponentially decreasing rate. Subsequently, we set a clear range of learning rates by cyclically vary the rates from a pre-defined range from $2 \times 10^{-7}$ to $2 \times 10^{-5}$. While multiple function forms can vary the rate cyclically, the triangular form (linearly increasing and then linearly decreasing) has demonstrated to be simple and effective for our problem in this paper.

*e). Architecture Fine-tuning and Compensating for Overfitting for Genus classification*: Having discussed choices for key hyperparameters in our architecture, we now present how our architecture is trained and fine-tuned for genus classification. Many steps are involved here including decision on which layer to start from the Inception-ResNet-V2 (IRV2) [25] architecture (among the 782 layers), how to add remaining layers for our specific genus classification problem, how to assign weights to them, how to avoid overfitting problems, and finally, when to stop the training.

For our problem, we initially started at layer 350 in the Inception-ResNet-V2 (IRV2) [25] architecture. Starting from layers too early will lead to poor learning, and starting from layers too deep will likely lead to over-fitting and also induce computational overhead. After initialization, subsequent layers were initialized with Glorot uniform initialization technique [9]. Since, IRV2 weights are highly optimized we didn't want to change them too much but wanted to optimize the weights of remaining layers. As such, we divided the training process in two phases. In first phase, we froze the weights of IRV2 layers and only allowed the changes in dense layers which we add. which we iteratively added with high learning rate. After 500 epochs, the loss curve reached the plateau where we saved the weights of the model. In second phase, we unfroze the weights of IRV2 and decreased the learning rate to 0.00001 to setup slow training. In this phase, every single weight across all layers in the architecture was allowed to change. After 1200 epochs, training and validation loss function attained a plateau and we stopped the training. Note that during training (and as is common in complex classification problems), we identified that our architecture suffered from overfitting problems, which we compensate by infusing a combination of three different regularization techniques between layers, namely dropout, early stopping and batch normalization [24], [13], [28].

*f). The Finalized Architecture for Genus Classification:* Table 2 illustrates the key parameters of the finalized architecture for classifying genus type of mosquitoes. The term *block*17_10_*Conv* denotes the $433^{rd}$ layer of the IRV2 architecture [25], upto which was utilized for our problem. The remaining layers (elaborated upon below) are added after that as specified in Table 2 for genus identification. The entries in the fields "Size In" and "Size Out" in the table refer to dimensions of the input and output matrices in the corresponding layer. After the $433^{rd}$ layer (of the IRV2 architecture), we add a Global Average Pooling Layer to reduce dimensionality to only one dimension, following which four dense layers are added to the architecture. Then, we concatenat the features of dense layers to preserve more information of each layer and finally perform softmax operation to output probabilities of classification between 0 and 1 for each of the three genera we attempt to classify (i.e., *Aedes*, *Anopheles* and *Culex*). This essentially summarizes our neural net architecture for genus classification.

## 4.3 Adaptations of our Architecture for other Classification Problems

Tables 3, 4 and 5 illustrate the architectures for classifying species within the *Aedes*, *Anopheles* and *Culex* genus types respectively. Recall again that these architectures are designed to classify a species within the corresponding genus types only. Finally in Table 6, we present the architecture for classifying directly the species only. Interestingly, the optimized architecture for this problem is the same one that was optimized for genus level classification, except that the number of possible classes here is *nine* (for the nine species), instead of *three* for the genus classification problem.

Table 2: Genus Architecture

| Layer | Size In | Size Out |
|---|---|---|
| $block17\_10\_conv$ (Layer 433 in IRV2) | $(None, 17, 17, 384)$ | $(None, 17, 17, 1088)$ |
| $GlobalAveragePooling$ | $(None, 17, 17, 1088)$ | $(1, 1088)$ |
| $dense\_1$ | $(1, 1088)$ | 512 |
| $dense\_2$ | 512 | 256 |
| $dense\_3$ | 256 | 128 |
| $dense\_4$ | 128 | 256 |
| $concat\_1$ | ($dense\_1$, $dense\_2$, $dense\_3$, $dense\_4$) | 1152 |
| $softmax$ | 1152 | 3 |

Table 3: Aedes Architecture

| Layer | Size In | Size Out |
|---|---|---|
| $conv2d\_93$ (Layer 346 in IRV2) | $(None, 17, 17, 160)$ | $(None, 17, 17, 192)$ |
| $GlobalAveragePooling$ | $(None, 17, 17, 192)$ | $(1, 192)$ |
| $dense\_1$ | $(1, 192)$ | 512 |
| $dense\_2$ | 512 | 512 |
| $dense\_3$ | 512 | 256 |
| $dense\_4$ | 256 | 128 |
| $concat\_1$ | ($dense\_1$, $dense\_4$) | 640 |
| $softmax$ | 640 | 3 |

Table 4: Anopheles Architecture

| Layer | Size In | Size Out |
|---|---|---|
| $block17\_8\_conv$ (Layer 401 in IRV2) | $(None, 17, 17, 384)$ | $(None, 17, 17, 1088)$ |
| $GlobalAveragePooling$ | $(None, 17, 17, 1088)$ | $(1, 1088)$ |
| $dense\_1$ | $(1, 1088)$ | 512 |
| $dense\_2$ | 512 | 512 |
| $dense\_3$ | 512 | 256 |
| $dense\_4$ | 256 | 256 |
| $dense\_5$ | 256 | 256 |
| $softmax$ | 256 | 3 |

Table 5: Culex Architecture

| Layer | Size In | Size Out |
|---|---|---|
| $conv2d\_111$ (Layer 407 in IRV2) | $(None, 17, 17, 128)$ | $(None, 17, 17, 160)$ |
| $GlobalAveragePooling$ | $(None, 17, 17, 160)$ | $(1, 160)$ |
| $dense\_1$ | $(1, 160)$ | 512 |
| $dense\_2$ | 512 | 128 |
| $dense\_3$ | 128 | 256 |
| $dense\_4$ | 256 | 512 |
| $dense\_5$ | 256 | 256 |
| $concat\_1$ | ($dense\_1$, $dense\_2$, $dense\_3$, $dense\_4$, $dense\_5$) | 2484 |
| $softmax$ | 2484 | 3 |

Table 6: Species-only Architecture

| Layer | Size In | Size Out |
|---|---|---|
| $block17\_10\_conv$ (Layer 433 in IRV2) | $(None, 17, 17, 384)$ | $(None, 17, 17, 1088)$ |
| $GlobalAveragePooling$ | $(None, 17, 17, 1088)$ | $(1, 1088)$ |
| $dense\_1$ | $(1, 1088)$ | 512 |
| $dense\_2$ | 512 | 256 |
| $dense\_3$ | 256 | 128 |
| $dense\_4$ | 128 | 256 |
| $concat\_1$ | ($dense\_1$, $dense\_2$, $dense\_3$, $dense\_4$) | 1152 |
| $softmax$ | 1152 | 9 |

## 5 EVALUATION OF OUR ARCHITECTURES

We now evaluate our neural net architectures for classifying mosquitoes from smartphone images using three strategies. The first is based on visualization of last convolution layer feature map, and the next one is based on accuracy of classification. We finally compare our AI based results with decades of existing research on mosquito phylogeny based on millions of years of their evolution.

### 5.1 Visualization of Feature Maps

Our first evaluation approach is to determine whether or not our architectures are actually able to focus in on the pixels corresponding to anatomical components within the mosquito image to classify, while simultaneously being able to exclude the background pixels.

To do so, once an input image is classified after the end of the Softmax layer, we traverse back in the architecture to identify the feature map for that particular image (at the conclusion of the last convolutional layer), and the weights corresponding to the class (i.e., the type of mosquito) that the image was identified with. If we denote the feature map for an image at Kernel $k$ in the last convolutional layer as $f_k(i, j)$, and if we denote the weight of each kernel connection for class $c$ as $w_k^c$, then, we compute the expression,

$$M_c(i, j) = \sum_k w_k^c f_k(i, j), \quad (1)$$

for each spatial location $(i, j)$ in the convoluted image. This essentially computes the importance of each feature map in the convoluted image when a classification is made. Subsequently, the value of $M_c(i, j)$ in the convoluted image is projected back onto the corresponding pixels in the original image to create a heatmap.

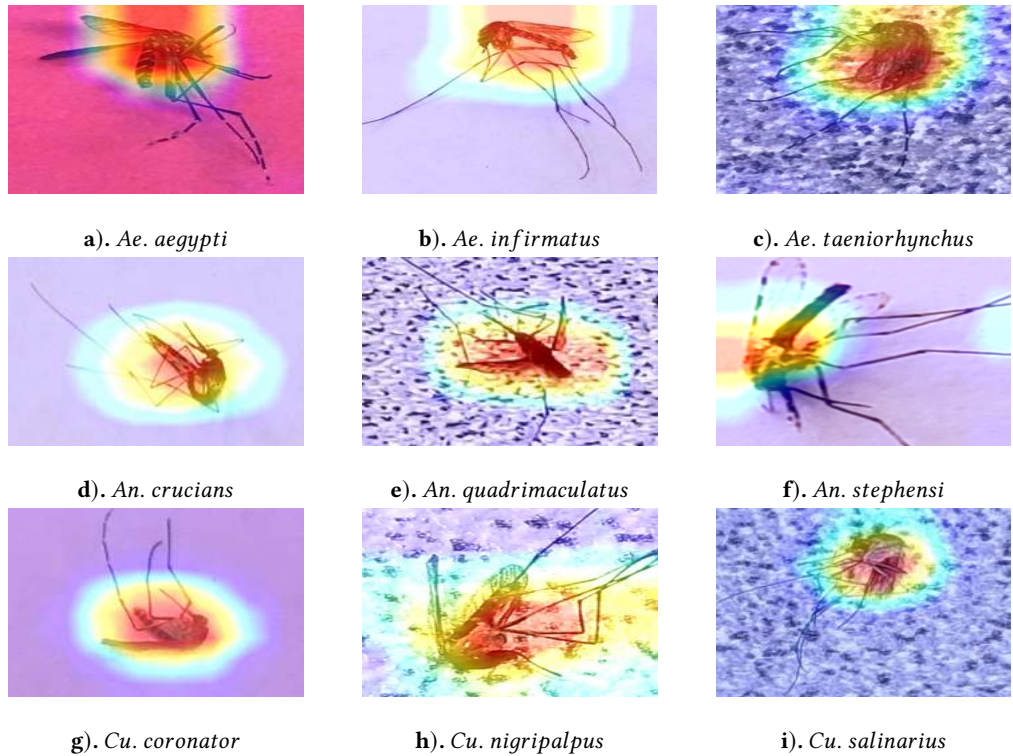

**Figure 2: One representative image sample with visualization of last layer convolution feature map in our dataset for each species classified, across multiple backgrounds and phones. This figure is best viewed in color.**

The higher the value of $M_c(i, j)$, the warmer the color of those corresponding pixels in the heatmap. These are the pixels within the image that were used predominantly to make the classification. As such, if in our images if most of the higher intensities are concentrated in critical anatomical components of the mosquito image, then we can trust our model better, as elaborated below.

**Discussions on the Results:** In Figure 2, we specifically highlight one representative feature map image for each of the nine species (for the architecture in Table 6) we attempt to classify, but our results are generalizable. We see from Figure 2 that irrespective of the background, or phone, the pixels with the highest intensities are concentrated in the most critical anatomical components of the image (thorax, scutum, wings, abdomen and legs). In fact, these are the anatomical components that are most vital for visual classification of a mosquito specimen - thorax and legs for *Ae. aegypti*; wings for *An. crucians* and *An. quadrimaculatus*; abdomen for *Ae. taeniorhynchus*; thorax for *Cu. nigripalpus* and scutum for *An. stephensi*. The taxonomists who saw these results were convinced of the fidelity of our AI approach to classify mosquito vectors.

## 5.2 Results on Classification Accuracies

We now present our results on accuracy of classification. We reiterate that our image dataset of 6807 images generated from 250 mosquito specimens presented in Table 1. Subsequently, 30% of these images (2042 in number) were separated out for validation alone. The remaining 4765 images were augmented for training (following procedures described in Section 3) to yield a dataset of 23825 images that were used to train and validate the architecture. The accuracy metric used to present results is Recall.

**Validation Accuracy during Training:** Table 7 presents results for the three problems attempted in this paper - genus only, species within genus, and species only. The classification accuracies presented are those wherein the accuracy of classification with the training images (23825 images) best matches the classification of accuracy with validation images (2042) when training was concluded. The classification accuracies presented represent the proportion of correctly identified images within each class.

**Validation with Unseen Images:** To test with unseen images, we trapped another 50 mosquito specimens evenly distributed among the nine species for which the model was trained. For this, we define a term called as a *set* for each mosquito specimen, which consists of three images for that specimen captured with the same smartphone camera, and on top of the same background, and in three different camera orientations. For each of the 50 mosquito specimens, we generated 12 such sets using four different phones and on top of three different backgrounds (wherein no two sets per mosquito specimen have the same phone-background configuration). A total of 600 such sets were generated for testing, wherein we reiterate that each set had had a total of three images per mosquito specimen in different orientations, that were taken via the same phone, and on top of the same background. This dataset is completely unseen by our architectures. Classification results for this dataset are presented in Table 8. Note here that the identification of a mosquito specimen for each *set* (comprising of three images per specimen) is based on computing the maximum of the average of the class probabilities as outputted by our architecture for each of the three images in that set. The results are elaborated below.

Table 7: Validation Accuracy

| Genus | % of Accuracy | Species Within Genus | % of Accuracy | Species | % of Accuracy |
|---|---|---|---|---|---|
| Aedes | 92% | aegypti | 85% | aegypti | 86% |
| | | infirmatus | 84% | infirmatus | 83% |
| | | taeniorhynchus | 81% | taeniorhynchus | 78% |
| Anopheles | 93.5% | crucians | 89% | crucians | 93% |
| | | quadrimaculatus | 79% | quadrimaculatus | 72% |
| | | stephensi | 98% | stephensi | 100% |
| Culex | 92% | coronator | 69% | coronator | 68% |
| | | nigripalpus | 64% | nigripalpus | 71% |
| | | salinarius | 72% | salinarius | 63% |

Table 8: Testing Set Accuracy

| Genus | % of Accuracy | Species Within Genus | % of Accuracy | Species | % of Accuracy |
|---|---|---|---|---|---|
| Aedes | 81% | aegypti | 82% | aegypti | 82% |
| | | infirmatus | 90% | infirmatus | 40% |
| | | taeniorhynchus | 50% | taeniorhynchus | 38% |
| Anopheles | 77% | crucians | 89% | crucians | 73% |
| | | quadrimaculatus | 60% | quadrimaculatus | 40% |
| | | stephensi | 98% | stephensi | 93% |
| Culex | 92% | coronator | 48% | coronator | 20% |
| | | nigripalpus | 43% | nigripalpus | 40% |
| | | salinarius | 51% | salinarius | 30% |

**Discussions on Results:** First, off the accuracy of classification during model training and validation in Table 7 is better than those with testing on unseen data after training in Table 8. This is natural, although, the accuracies are not too far off, hence justifying the fidelity of our model. Secondly, we see that genus classification accuracies are high. This is also encouraging since identifying the genus itself is very useful during outbreaks. Once we start the finer-grained process of identifying species, we see interesting results. Mosquitoes belonging to the *Aedes* genus are being classified better. It is generally true that even for experts, visually recognizing *Ae. aegypti* (a deadly vector for many diseases) is easy because of the very distinct 'lyre' shaped pattern on their thorax. Our model can also identify *An. crucians* and *An. stephensi* much more accurately. This is because *An. crucians* mosquitoes have three distinct dark spots on the wings, which aid identification. However, *An. quadrimaculatus* mosquitoes have four dark spots on their wings, and so there is some confusion between *An. crucians* and *An. quadrimaculatus* mosquitoes. *An. stephensi* mosquitoes have distinct yellow colored scutum, thorax and wings, which makes it a species easy to identify. Most confusion happens with species belonging to the *Culex* genera, since they share similar brownish-copper color patches with each other, and also with *Ae. infirmatus* mosquitoes. The corresponding confusion matrix is shown in Figure 3b. Taxonomists we partnered with again agreed on the validity of our results, and confusions among the classes, as elaborated in the next subsection below.

### 5.3 Comparing our Results with Mosquito Phylogeny based on Evolution

The classification accuracies and confusion we see in our AI models match the evolutionary relationships of mosquitoes across millions of years. These relationships are reflected in the phylogenetic tree Figure 3a, primarily reconstructed based on genetic sequencing [10] [11] [22]. Starting from the left, the *Anopheles* clade diverged from the other mosquitoes ∼ 217 million years (mya) ago [22]. Interestingly, the *Anopheles* mosquitoes here have the highest classification accuracies in our model, and within the *Anopheles* genus, the Old World species, *An. stephensi* (∼ 107 mya divergence) [19] has a higher classification accuracy in our model than those of the more closely-related New World sister groups, *An. crucians* and *An. quadrimaculatus*. Similarly, within *Aedes*, *Ae. aegypti* is the most evolutionarily distinct species (∼ 92 mya divergence), and it has the highest classification accuracy in our model for the *Aedes* genus. The evolutionary relationships among species in the *Culex* genus were unable to be resolved morphologically even by biologists today (illustrated pictorially in Figure 3a) [11], and this fact is indeed reflected in the lower levels of classification accuracy in our model for species in the *Culex* genus (Figure 3b).

Ultimately, this suggests that evolution – as measured by relative divergence times – drives greater anatomical disparity, which in turn yields higher classification accuracies when using AI. In other words, our paper shows that signals from millions of years of natural selection is being revealed by real-time AI. Naturally, these discoveries provide much more validity to our work in this paper, and also its potential for future broader impact in diverse scientific disciplines, apart from automated mosquito vector surveillance.

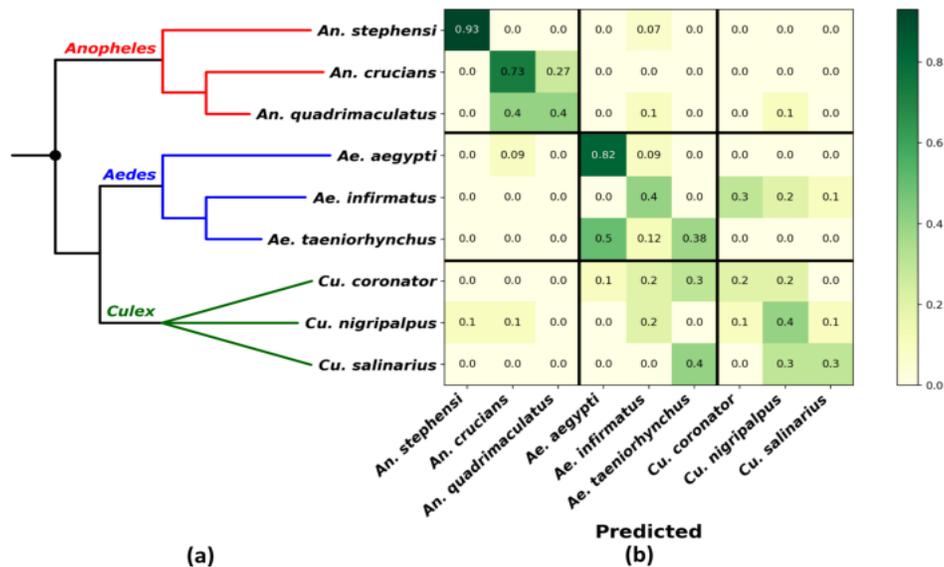

Figure 3: (a) Phlyogenetic tree showing evolutionary relationships of the mosquitoes [10] [11] [22]. Leftward is further back in geologic time. Black dot denotes the common ancestor, ∼ 217 million years ago [22]. (b) Species testing set confusion matrix with classification accuracies. This Figure is best viewed in Color.

## 6 DISCUSSIONS ON PRACTICAL IMPACT

First off, identification of trapped mosquitoes is a critical component of public health efforts. However, this process today is manual, time-consuming and cognitively demanding due to long hours spent peering through a microscope. The taxonomists we partnered with in India, Sub-Saharan Africa, Brazil and USA are all close to 70 yrs old, and they indicated that taxonomy is indeed a dying field, and related expertise is hard to attract and train. Our approach can automate classification wherein after trapping, mosquitoes bought to the lab could be scattered on a plain flat board (under normal light conditions). A movable camera could possibly hover over each mosquito, autofocus, take a few pictures from multiple angles, upload results to a cloud or a local server (where classification and storage happens). Images of critical vectors when identified by AI can be immediately sent to taxonomists via a dahsboard for further inspection. We are already engaging with our public health partners to pilot such a system. Our partners were excited about a) improved operational efficiency of automated surveillance; b) high accuracies in classifying *Ae. aegypti* and *An. stephensi*; c) robustness of our feature maps in learning from critical anatomical components of mosquitoes.

Secondly, citizens with their smartphones, could capture and share images of mosquitoes they encounter. Such data could help generate superior real-time distribution maps of disease vectors. Remarkably, one major effort in this realm is the iNaturalist project, where citizens are encouraged to capture and upload images of mosquitoes they encounter in nature. Our work in this paper can add value to such efforts on a global scale.

It is pertinent to mention that our datasets are limited to nine vector species in Hillsborough county (in Florida) only. Expanding our dataset geographically is part of our on-going work.

## 7 CONCLUSIONS

In this paper, we design deep neural network architectures to classify the genus and species of mosquitoes from smartphone images. Our architectural framework employs the principle of Transfer Learning, and is a hybrid of Inception Net and ResNet architectures. We believe that our results, and the right contexts in which the results should be interpreted give us the confidence that our proposed system is practical. We are aware of recent works in the space of fine-grained object recognition using neural networks [29] [7], that are also approaches to improve mosquito classification accuracy, and is part of on-going work. Currently, we are focusing on piloting our system in low income countries where taxonomic expertise is harder to find. Should systems like ours be adopted, and resulting image datasets shared, we can expect significantly more efficient mosquito surveillance programs across the globe.